\pdfoutput=1
\documentclass[11pt]{article}

\usepackage{acl}
\usepackage{lipsum}
\usepackage{times}
\usepackage{latexsym}
\usepackage{subcaption}
\usepackage{multirow}
\usepackage{xcolor}
\usepackage[T1]{fontenc}
\usepackage{setspace}
\usepackage{multicol}
\usepackage{tikz}
\usepackage[utf8]{inputenc}

\usepackage{booktabs}
\usepackage{graphicx}
\usepackage{microtype}

\usepackage{inconsolata}
\usepackage[most]{tcolorbox}
\title{Unveiling Divergent Inductive Biases of LLMs on Temporal Data}

\author{Author 1  \and Author n \\
        Address line \\ Address line}
\author{Sindhu Kishore \\
  University of Rochester \\
  \texttt{skishor2@ur.rochester.edu} \\\And
   Hangfeng He \\
  University of Rochester \\
  \texttt{hangfeng.he@rochester.edu} \\}

\begin{document}

\maketitle
\begin{abstract}

Unraveling the intricate details of events in natural language necessitates a subtle understanding of temporal dynamics. Despite the adeptness of Large Language Models (LLMs) in discerning patterns and relationships from data, their inherent comprehension of temporal dynamics remains a formidable challenge. This research meticulously explores these intrinsic challenges within LLMs, with a specific emphasis on evaluating the performance of GPT-3.5 and GPT-4 models in the analysis of temporal data. Employing two distinct prompt types, namely Question Answering (QA) format and Textual Entailment (TE) format, our analysis probes into both implicit and explicit events. The findings underscore noteworthy trends, revealing disparities in the performance of GPT-3.5 and GPT-4. Notably, biases toward specific temporal relationships come to light, with GPT-3.5 demonstrating a preference for ``AFTER'' in the QA format for both implicit and explicit events, while GPT-4 leans towards ``BEFORE''. Furthermore, a consistent pattern surfaces wherein GPT-3.5 tends towards ``TRUE'', and GPT-4 exhibits a preference for ``FALSE'' in the TE format for both implicit and explicit events. This persistent discrepancy between GPT-3.5 and GPT-4 in handling temporal data highlights the intricate nature of inductive bias in LLMs, suggesting that the evolution of these models may not merely mitigate bias but may introduce new layers of complexity.\footnote{Our code is publicly available at \url{https://github.com/SindhuKRao/LLM_temporal_Bias}.}.

% \footnote{Our code is publicly available at \href{https://github.com/SindhuKRao/LLM_temporal_Bias.}}
%These findings contribute to our comprehension of temporal challenges in LLMs, providing valuable insights into the advancements and biases inherent in GPT-3.5 and GPT-4.
% In recent years, Large Language Models (LLMs) have attracted substantial attention due to their capabilities in producing human-like text, responding to queries, and performing translations, exemplified by models such as ChatGPT, GPT-3.5 and GPT-4. While these LLMs offer numerous advantages, they also raise concerns related to ethics, bias, responsible AI development, and potential technology misuse. One persistent challenge for LLMs has been accurately interpreting temporal data. This research focuses on investigating biases inherent in LLMs when analyzing temporal data, with a dedicated scrutiny on the performance of GPT-3.5 and GPT-4 models. The aim is to uncover inconsistencies and biases observed in their treatment of temporal data. By comparing the performance of each model, we seek to determine whether newer iterations address inherent issues found in their predecessors.Our analysis delves into implicit and explicit events, examining two different prompt types: Question-Answer format and Textual format. This multifaceted approach allows us to understand how the models handle varying levels of complexity. Through this study, we contribute to ongoing discussions regarding ethical AI development by exploring these complexities, emphasizing the importance of mitigating biases in LLMs for accurate temporal analysis.
\end{abstract}

\section{Introduction}

Temporal relations play a crucial role across diverse applications, including event summarization \citep{10.1007/s11280-017-0501-x, keith2023survey}, predicting future events \citep{future-event}, and medical information processing \citep{jung-etal-2011-building, alfattni2020extraction}. Despite their importance, LLMs, especially those with limited context windows, face challenges in accurately sequencing events due to intricate temporal dependencies. Efforts have been devoted to developing methodologies for effective temporal relation extraction \citep{choubey-huang-2017-sequential, ning2018cogcomptime, ning2019improved, wang-etal-2020-joint, zhang-etal-2022-extracting}, along with initiatives to create benchmark datasets with a temporal focus \citep{TBcorpus, verhagen-etal-2010-Tempeval2, ning2018multi, zhou2021temporal, gantt2022decomposing}. %However, the complex task of discerning causal relationships between events adds further challenges, potentially leading to misunderstandings.
However, discerning causal relationships between events adds complexity and can lead to misunderstandings. This complexity is strengthened by the absence of explicit temporal reasoning mechanisms, introducing biases in models' predictions and preferences for specific temporal relations.
% \vspace{-5pt}
\begin{figure}[!t]
  \centering
\begin{subfigure}{0.8
\linewidth}
    \includegraphics[width=5cm]{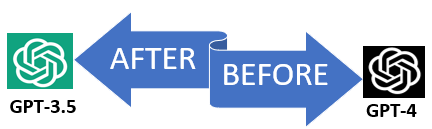}
    \label{subfig:1}
  \end{subfigure}
  \hfill
    \centering
  \begin{subfigure}{0.8\linewidth}
    \includegraphics[width=5cm]{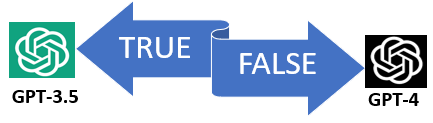}
    \label{subfig:2}
  \end{subfigure}
  \vspace{0pt}
  \caption{ Inductive bias in OpenAI LLMs: GPT-4 exhibits a preference for BEFORE and FALSE, while GPT-3.5 tends to favor AFTER and TRUE.}
  \label{fig:main}
\end{figure}
\vspace{-2pt}
Surprisingly, a notable gap exists in research exploring inductive bias in LLMs when discerning temporal relations. Our study investigates the temporal comprehension abilities of GPT-3.5 and GPT-4 \citep{OPENAI}, aiming to understand their grasp of temporal relationships. Despite frequent model updates, significant biases were unveiled. Using Question Answering (QA) and Textual Entailment (TE) prompts, we queried both models to determine temporal relations. Illustrated in Figure \ref{fig:main}, the results expose variations in GPT-3.5 and GPT-4 performance, revealing biases towards specific temporal relationships. GPT-3.5 favors "AFTER" in QA for implicit and explicit events, while GPT-4 leans towards "BEFORE." In TE, GPT-3.5 tends towards "TRUE," and GPT-4 prefers "FALSE" for both implicit and explicit events.

\section{Methodology}
\label{sec:methodology}

Our analysis involved two %distinct datasets 
types of temporal data: one focusing on implicit events, actions, or situations not directly articulated in the text but inferred from context, while the other centered on explicit events explicitly mentioned in the context. Furthermore, we delved into two prompt formats to gauge their influence on aiding LLM %comprehension and precision 
in generating responses. These formats comprised the QA format, where questions prompt the model, and the TE format, tailored to assess the logical relationships within sentences specifying temporal relations.

\paragraph{Question answering format.}We initially conducted experiments using the QA format, focusing on explicit events. In this configuration, we tasked both models with determining the temporal relation (``BEFORE'' or ``AFTER'') between two provided events within the given context. The same approach was applied to implicit events. %Fig 2
Figure \ref{fig:examples} provides the template and examples illustrating this format.
% Next, we curated the implicit events TRACIE\citep{zhou2021temporal}, simplifying the relations by consolidating "starts after/ends after" as "after" and "starts before/ends before" as "before". As the dataset was structured in the Entailment format, we extracted pairs of two events labeled as event 1 and event 2, accompanied by their corresponding relations. It's worth noting that the TRACIE dataset primarily focuses on implicit events. Subsequently, we converted the TRACIE dataset into the QA format, using this structure to prompt both models for assessments of temporal relations.Examples illustrating this format are in Table 1.
% We selected events from the TimeBank\citep{TBcorpus}, Tempeval\citep{verhagen-etal-2010-Tempeval2}, and AQUAINT datasets based on their relations, specifically focusing on those categorized as AFTER/IAFTER or BEFORE/IBEFORE. 
\begin{figure*}[t] 
\begin{tcolorbox}[colframe=orange!75!black, title=Dataset, fonttitle=\bfseries\small, width=\textwidth, halign title=center,notitle, before upper=\small]
\centering\textbf{QA:Implicit Event}\\
\raggedright
\textbf{Template}:\textit{ \textcolor{purple}{context:<context>.E1:<event1>.E2:<event2>.
For the given context, give the temporal relation
between E1 and E2 as before or after. Give your
answer in the format, Relation:<ans>.}}\\
\textbf{Prompt}:\textcolor{blue}{Context:'Trisha was a manager of a local diner. It was a slow night. The diner hadn’t been making any money.
Trisha didn’t want the diner to lose any more money. Trisha closed the diner an hour early.'E1:'Trisha got a job at a diner '.
E2:'Trisha closed the diner'.}
For the given context, give the temporal relation between E1 and E2 as before or after. Give your answer in the format, Relation:<ans>.\\
\raggedright
\textbf{Response}:Relation:AFTER.\\
% \vspace{0pt}
%     \begin{tikzpicture}
%       \draw[dashed, orange!75!black] (0,0) -- (\textwidth,0);
%     \end{tikzpicture}
% \vspace{2pt}
\centering\textbf{QA:Explicit Event}\\
\raggedright
\textbf{Template}:\textit{ \textcolor{purple}{Context:<context>.For the given context,give the temporal relation between (eid:<event>) and (eid:<event>) as BEFORE or AFTER. Give your answer in the format, Relation:<ans>. }}\\
\textbf{Prompt}:\textcolor{blue}{context:'The American ambassador to Kenya was among hundreds (e12:injured), a local TV (e4:said)'}.For the given context, give the temporal relation between (e12:injured) and (e4:said) as before or after. Give your answer in the format, Relation:<ans>.\\
% \textbf{Prompt}:Context:'Crown Leasing KK (e3:applied) to be (e4:wound) up at the Tokyo District Court with two other Nippon Credit Bank Ltd (NCB)'.For the given context, give the temporal relation between (e3:applied) and (e4:wound) as BEFORE or AFTER. Give your answer in the format , Relation:<ans>\\
\raggedright
\textbf{Response}:Relation:BEFORE.\\
% \vspace{0pt}
%     \begin{tikzpicture}
%       \draw[dashed, orange!75!black] (0,0) -- (\textwidth,0);
%     \end{tikzpicture}
% \vspace{2pt}
\centering\textbf{TE:Implicit Event}\\
\raggedright
\textbf{Template}:\textit{ \textcolor{purple}{context:<context>.For the given
context,state if the statement event:<event> is 
True or False. Give your answer in the format
, Ans:<ans>.}}\\
\textbf{Prompt}:\textcolor{blue}{context:'Trisha was a manager of a local diner. It was a slow night. The diner hadn't been making any money. Trisha didn't want the diner to lose any more money. Trisha closed the diner an hour early.'}For the given context,state if the statement event:' Trisha got a job at a diner starts before Trisha closed the diner ' is True or False. Give your answer in the format , Ans:<ans>.\\
\raggedright
\textbf{Response}: Ans:True.\\
\textbf{Prompt}:\textcolor{blue}{context:'Trisha was a manager of a local diner. It was a slow night. The diner hadn't been making any money. Trisha didn't want the diner to lose any more money. Trisha closed the diner an hour early.'}.For the given context,state if the statement event:' Trisha got a job at a diner starts after Trisha closed the diner ' is True or False. Give your answer in the format , Ans:<ans>.\\
\textbf{Response}: Ans:False.\\
% \raggedright
% \textbf{Response}: Ans:False\\
% \vspace{0pt}
%     \begin{tikzpicture}
%       \draw[dashed, orange!75!black] (0,0) -- (\textwidth,0);
%     \end{tikzpicture}
% \vspace{2pt}
\centering\textbf{TE:Explicit Event}\\
\raggedright
\textbf{Prompt}:\textcolor{blue}{context:’The American ambassador to Kenya was among hundreds (e12:injured), a local TV
(e4:said)’}.For the given context, state if the statement: 'event (e12:injured) is BEFORE (e4:said)’
is True or False. Give your answer in the format , Ans:<ans>.\\
\textbf{Response}: Ans:True.\\
\textbf{Prompt}:\textcolor{blue}{context:’The American ambassador to Kenya was among hundreds (e12:injured), a local TV
(e4:said)’}.For the given context,state if the statement: 'event (e12:injured) is AFTER (e4:said)’ is
True or False. Give your answer in the format , Ans:<ans>.\\
\textbf{Response}: Ans:False.
\end{tcolorbox}
\caption{\textcolor{purple}{Template }and \textcolor{blue}{Examples} of QA and TE prompts for implicit \& explicit events.}
\label{fig:examples}
\end{figure*}

\paragraph{Textual Entailment Format.}
Subsequently, we employed the textual entailment format as the next prompt type. In this format, we presented the model with a context along with a sentence declaring the temporal relation between two events, and then tasked the model with assessing its truthfulness. For every pair, there exists one TRUE and one FALSE label, as one corresponds to the gold label, and the other represents an incorrect label.Examples illustrating this format are provided in Figure \ref{fig:examples}. 
% we generated a set containing both the gold relation as 'TRUE' and its inverse as 'FALSE', labeling them as true and false, respectively. 
% The TRACIE dataset were obtained in a textual entailment format. We utilized this dataset as an alternative format to evaluate both models' performance regarding temporal comprehension.We transformed the labels ,positives to 'True' and negatives to 'False'. We then used this Textual Entailament Format to conduct our analysis on the TimeBank series. We combined the 2 events within the context with its relation and inverse relation to create 2 pairs and labeled them as True or False based on their actual Relationship.
% Examples for this format can be found in Table 2.

\paragraph{Inductive Bias Measurement:}
In our evaluation, we focused on probing the model's inductive biases related to temporal relations. %Inductive bias refers to the inherent inclinations or limitations within a model's decision-making processes, influencing its learning patterns and preferences for specific relationships. 
To quantify the inductive bias, we examined the model's preference for ``BEFORE'' and ``AFTER'' relations in the QA format and assessed its tendencies toward ``TRUE'' and ``FALSE'' in the Textual Entailment format.

\section{Experimental Settings}
\label{sec:experiments}

\paragraph{Dataset}
For our experimentation, we employed datasets such as TimeBank \citep{TBcorpus}, Tempeval \citep{verhagen-etal-2010-Tempeval2}, AQUAINT, and TRACIE \citep{zhou2021temporal}. The analysis of implicit and explicit events was conducted separately. We extracted "BEFORE"/"IBEFORE" as "BEFORE" and "AFTER," /"IAFTER" as "AFTER" from TimeBank, events from Task C in TempEval featuring "BEFORE" or "AFTER" relations. This yielded 1576 explicit events from TimeBank, TempEval, and AQUAINT datasets, comprising 815 "AFTER" and 761 "BEFORE" events. The dataset was duplicated for the Entailment format, creating inverse relations with gold as "TRUE" and inverse as "FALSE," expanding the dataset to 3150 events with 1575 "TRUE" and "FALSE" values.

For Implicit events, the TRACIE dataset \citep{zhou2021temporal} was used. Transforming "starts after/ends after" into "AFTER" and "starts before/ends before" into "BEFORE”,the dataset included a total of 22,050 events evenly distributed between "TRUE" and "FALSE" labels, representing gold and inverse relations. Among the 11,025 gold relations, 4,659 were identified as "AFTER," while 6,366 were classified as "BEFORE".

\paragraph{Large language models.}

The GPT series, renowned as the leading range of Large Language Models (LLMs), holds widespread popularity. Our analysis began with these models due to their extensive usage, leaving the investigation of biases in other LLMs for potential future research. We conducted our analysis using OpenAI's GPT-3.5 and GPT-4 models, specifically employing the latest stable versions: gpt-3.5-turbo-1106 and gpt-4-1106-preview. The gpt-3.5-turbo-1106 model has a default context window of 16k tokens, while GPT-4 features a context window of 128k tokens. 
% This improvement enhances the model's responses by ensuring better continuity and coherence. The knowledge available in GPT-3.5-turbo-1106 extends up to September 2021. On the other hand, GPT-4 Turbo possesses advanced capabilities, encompassing knowledge of global events until April 2023 and featuring a 128k context window, enabling it to incorporate over 300 pages of text within a single prompt.

\paragraph{Performance Measurement}
We assessed bias by examining patterns in prediction preferences, aiming to determine if the models consistently favored or exhibited imbalances in predicting specific temporal relations. %To gauge the models' performance, we measured the disparities between the predicted and actual distribution of relations in the dataset. 
We scrutinized tendencies towards ``BEFORE'' and ``AFTER'' relations in the QA format, and in the %Entailment 
TE format, we analyzed biases towards ``TRUE'' and ``FALSE''. Furthermore, we tested the models for consistency by presenting identical events and contexts with reversed temporal relations.

\section{Results and Analysis}
\label{sec:results}
\begin{figure*}[!htp]
% \centering
% \textbf{Bias in QA format}\par\medskip
% \raggedcolumns
\begin{minipage}[b]{0.5\linewidth}
\includegraphics[width=7cm]{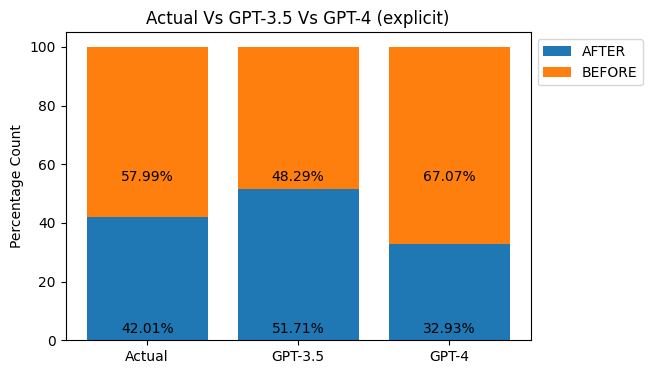}
% \caption{Comparison between Actual , GPT 3.5 and
% GPT 4 for explicit events}\label{label-a}
\end{minipage}\qquad
\begin{minipage}[b]{0.5\linewidth}
\includegraphics[width=7cm]{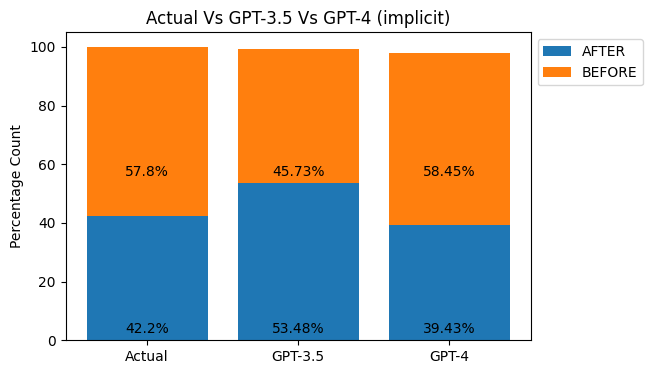}
% \caption{Comparison between Actual , GPT 3.5 and
% GPT 4 for explicit events}\label{label-b}
\end{minipage}
\caption{GPT-3.5 biased towards AFTER and GPT-4 biased towards BEFORE in QA.}
\label{fig:label-QA}
\end{figure*}

\paragraph{BEFORE/AFTER bias in QA:}
We evaluated the models' performance in the QA format for explicit events. Among 1576 instances, comprising 914 with a ``BEFORE'' relation and 662 with an ``AFTER'' relation, GPT-3.5 demonstrated a bias towards 815 prompts as ``AFTER'' and 761 as ``BEFORE'', indicating a preference for AFTER, as shown in Figure \ref{fig:label-QA}. In contrast, GPT-4 exhibited a preference for ``BEFORE'', leaning towards 1057 prompts as ``BEFORE'' and 519 as ``AFTER'', revealing a divergent pattern between the two models.%\\
% \vspace{5pt}

For implicit events, totaling approximately 11,652 entries, with 6,735 indicating a ``BEFORE'' relation and 4,917 an ``AFTER'' relation, both models displayed patterns resembling those observed in explicit events. GPT-3.5 predominantly favored the ``AFTER'' relation, identifying 6,232 instances as ``AFTER'', 5,329 as ``BEFORE'', and approximately 91 as indeterminable. Conversely, GPT-4 leaned towards ``BEFORE'', marking 6,811 instances as ``BEFORE'', 4,594 as ``AFTER'', and 247 as indeterminable. The contrasting outcomes between the models in both explicit and implicit events add an intriguing and contradictory dimension to their assessments.%\\
\vspace{2pt}
\paragraph{Consistency in TE.}We now analyzed the results of TE format. We encountered an unexpected pattern in both the implicit and explicit events. We had anticipated an even distribution of 'True' and 'False' responses due to the contradictory pairs as discussed in Section \ref{sec:methodology}. However, we found inconsistencies where the model consistently produced matching responses—yielding (``True'', ``True'') or (``False'', ``False'') rather than the expected mix of (``True'', ``False'') or (``False'', ``True'') in numerous instances. To delve deeper, we categorized our findings into consistent and inconsistent pairs for further examination.
\vspace{0pt}

\begin{figure*}[htb]

\begin{minipage}[b]{0.5\linewidth}
\includegraphics[width=7cm]{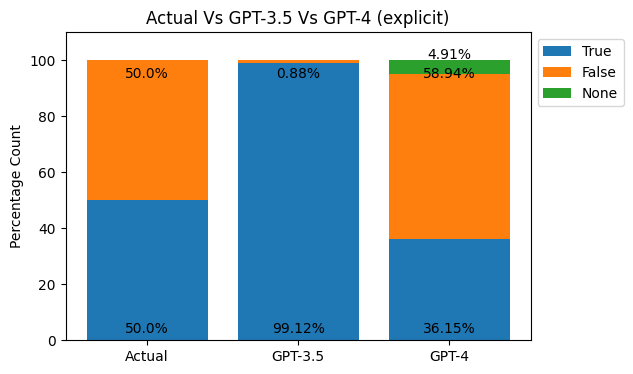}
\end{minipage}\qquad
\begin{minipage}[b]{0.5\linewidth}
\includegraphics[width=7cm]{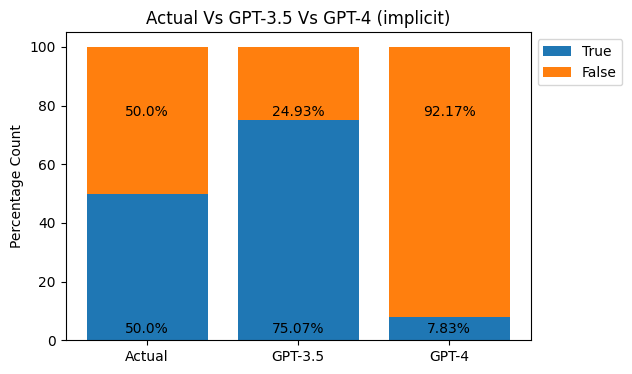}
\end{minipage}
\caption{GPT-3.5 biased towards TRUE and GPT-4 biased towards FALSE in TE -Inconsistent pair.}\label{label-TE}
\label{fig:label-TE}
\end{figure*}
\vspace{-10pt}

% \begin{figure*}[ht]
% \centering
% \begin{minipage}[b]{.4\textwidth}
% \includegraphics[width=8cm]{figures/consistency_TB.png}
% \caption{Consistency in Response for explicit events}\label{label-a}
% \end{minipage}\qquad
% \begin{minipage}[b]{.4\textwidth}
% \includegraphics[width=8cm]{figures/consistency_TRACIE.png}
% \caption{Consistency in Response for implicit events}\label{label-b}
% \end{minipage}
% \end{figure*}

\paragraph{TRUE/FALSE bias in TE-Inconsistent Pair}
These pairs contain actual values of ``True'' and ``False'', yet the predicted values consistently align as either (``True'', ``True'') or (``False'', ``False''). For implicit events, GPT-3.5 exhibited approximately 83.3\% inconsistency, while GPT-4 showed 67.1\% inconsistency. In explicit events, GPT-3.5 demonstrated 94.6\% inconsistency, whereas GPT-4 presented only 32.4\% inconsistent results. Comparing both models, it's evident that GPT-4 displays greater consistency compared to GPT-3.5 based on these findings. Upon further analysis of these inconsistent results to check for biases, we made a surprising discovery. GPT-3.5 tends to show a bias towards ``True'', while GPT-4 leans towards ``False'' as shown in Figure \ref{fig:label-TE}. This bias was consistently observed in both implicit and explicit events, revealing a contradicting bias between the models.
\vspace{-5pt}
\begin{table}[!htb]
\centering
  \fontsize{7pt}{7pt}\selectfont
  \begin{tabular}{|c|c|c|c|c|c|}
    \hline
    \rule{0pt}{8pt}
   \textbf{Model} & \textbf{Event} & \textbf{Relation} & \textbf{Actual} & \textbf{Prediction}\\
    \hline
     \rule{0pt}{8pt}
   \multirow{4}{*}{\textbf{GPT-3.5}} & \multirow{2}{*}{\textbf{Implicit}} & BEFORE & 48.02\% & 26.48\% \\
    & & AFTER & 51.98\% & \textcolor{blue}{73.52}\% \\
    \cline{2-5} % Add hline for the first three columns
    \rule{0pt}{8pt} % Adds space after the hline
    & \multirow{2}{*}{\textbf{Explicit}} & BEFORE & 50.00\% & 50.00\% \\
    & & AFTER & 50.00\% & 50.00\% \\
    \hline
     \rule{0pt}{8pt}
     \multirow{4}{*}{\textbf{GPT-4}} & \multirow{2}{*}{\textbf{Implicit}} & BEFORE & 62.82\% & \textcolor{blue}{70.42\%} \\
    & & AFTER & 37.18\% & 29.58\% \\
    \cline{2-5} % Add hline for the first three columns
    \rule{0pt}{8pt} % Adds space after the hline
    & \multirow{2}{*}{\textbf{Explicit}} & BEFORE & 50.00\% & 50.00\% \\
    & & AFTER & 50.00\% & 50.00\% \\
    \hline
  \end{tabular}
  
  \caption{Actual vs Predicted distribution in consistent TE.}
  % \caption{Actual vs Predicted distribution of GPT-3.5 and GPT-4 in consistent TE.}
  \label{tab:te-consis}
\end{table}
% \vspace{0pt}
% -------------------------
\vspace{-30pt}
\paragraph{BEFORE/AFTER bias in TE-Consistent Pair}
These pairs encompass actual values of ``True'' and ``False'', with predicted values aligning as either (``True'', ``False'') or (``False'', ``True''). For implicit events, roughly 16.7\% of GPT-3.5's total results were consistent, while GPT-4 showed 32.9\% consistency, while explicit events had, approximately 5.4\%  consistency for GPT-3.5 and 67.6\% consistency in GPT-4. Upon examining these consistent pairs to detect bias toward ``BEFORE'' and ``AFTER'', we noted a familiar pattern. For implicit events GPT-3.5 displayed a bias toward ``AFTER'', whereas GPT-4 leaned toward ``BEFORE''. However, the results were notably consistent and unbiased for explicit events as shown in Table\ref{tab:te-consis}. This discrepancy might arise since the model is able to comprehend context more effectively and provide unbiased predictions. In contrast, the implicit events poses challenges for the model to assess accurately, potentially leading to biased results. Additional information is available in Appendix \ref{sec:appendix}.
\begin{figure}[htbp]
    \centering
    \begin{subfigure}[b]{0.45\textwidth}
        \centering
%            \textbf{\begin{center}\small
%  Confusion Matrix for QA Format (GPT-3.5)
% \end{center} }\par\medskip
        \includegraphics[width=\linewidth]{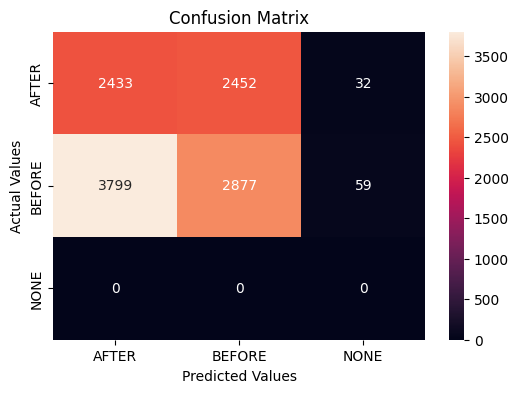} 
        \caption{GPT-3.5 : QA implicit events.}
        \label{fig:cfm_QA_GPT3_5_implicit}
    \end{subfigure}
    \hfill
    \begin{subfigure}[b]{0.45\textwidth}
        \centering
         \includegraphics[width=\linewidth]{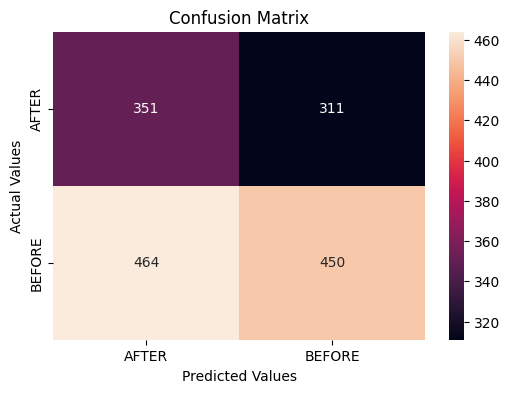}
  \caption{GPT-3.5 : QA explicit events.}
  \label{fig:cfm_QA_GPT3_5_explicit}
    \end{subfigure}
    \vskip\baselineskip
    \vspace{-5pt}
    % -------------------------------------------------------------
    \begin{subfigure}[b]{0.45\textwidth}
        \centering
%            \textbf{\begin{center}\small
%   Confusion Matrix for QA Format (GPT-4)
% \end{center} }\par\medskip
         \includegraphics[width=\linewidth]{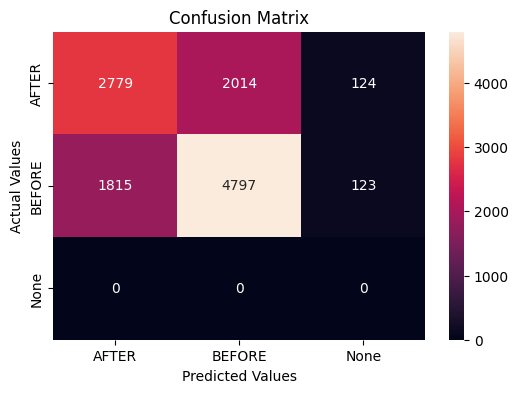}
  \caption{GPT-4 : QA implicit events.}
  \label{fig:cfm_QA_GPT4_implicit}
    \end{subfigure}
    \hfill
    \begin{subfigure}[b]{0.45\textwidth}
        \centering
         \includegraphics[width=\linewidth]{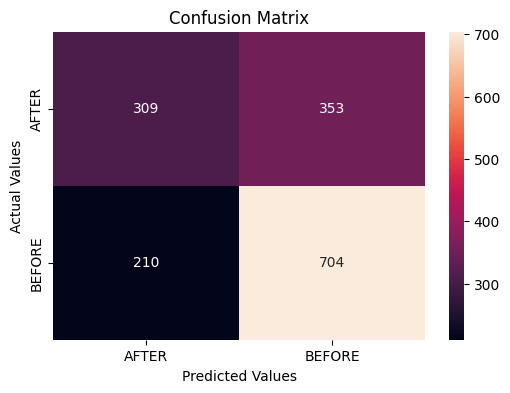}
  \caption{GPT-4 : QA explicit events.}
  \label{fig:cfm_QA_GPT4_explicit}
    \end{subfigure}
    \end{figure}
    % ----------------------------------------------------------
\begin{figure}[htbp]
    \centering
    \begin{subfigure}[b]{0.45\textwidth}
        \centering
 \includegraphics[width=\linewidth]{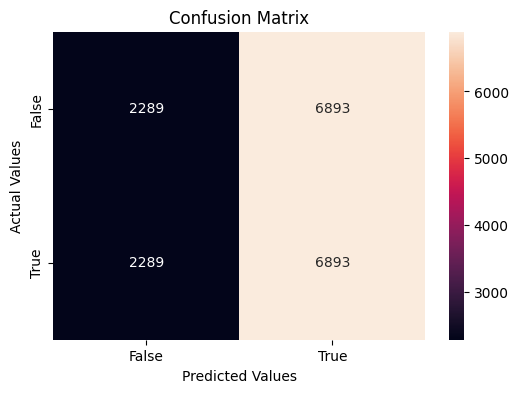}
  \caption{GPT-3.5 : inconsistent TE implicit events.}
  \label{fig:cfm_TE_GPT3_5_implicit}
    \end{subfigure}
    \hfill
    \begin{subfigure}[b]{0.45\textwidth}
        \centering
  \includegraphics[width=\linewidth]{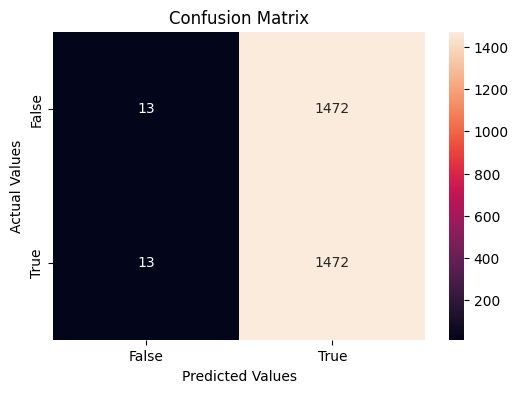}
  \caption{GPT-3.5 : inconsistent TE explicit events.}
  \label{fig:cfm_TE_GPT3_5_explicit}
    \end{subfigure}
    \vskip\baselineskip
    \vspace{-5pt}
    % --------------------------------------------------------
    \begin{subfigure}[b]{0.45\textwidth}
        \centering
  \includegraphics[width=\linewidth]{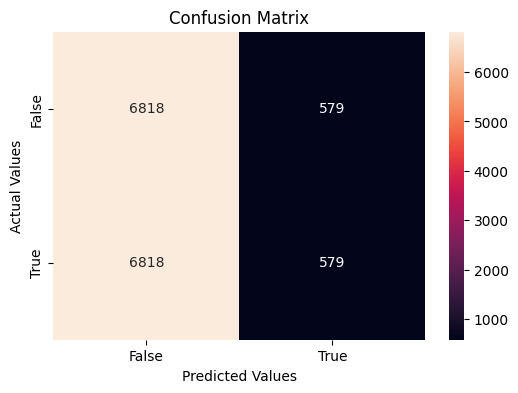}
  \caption{GPT-4 : inconsistent TE implicit events.}
  \label{fig:cfm_TE_GPT4_implicit}
    \end{subfigure}
    \hfill
    \begin{subfigure}[b]{0.45\textwidth}
        \centering
  \includegraphics[width=\linewidth]{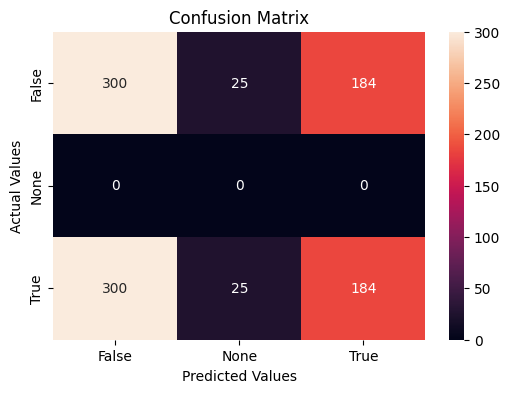}
  \caption{GPT-4 : inconsistent TE explicit events.}
  \label{fig:cfm_TE_GPT4_explicit}
    \end{subfigure}
    \end{figure}
\vspace{-20pt}

\section{Conclusion and Future Work}
Our study identified performance disparities between GPT-3.5 and GPT-4, with the latter showing more consistency. Notably, biases were observed, as GPT-3.5 favored ``AFTER'', while GPT-4 favored ``BEFORE'', and GPT-3.5 tended towards ``TRUE'', while GPT-4 favored ``FALSE''. This consistent yet contradictory pattern raises questions about whether new model releases might unintentionally introduce new biases. The observed biases across multiple datasets and prompt formats warrant a deeper exploration of the models' understanding of temporal data. %Although two different formats were used to assess prompt influence, no conclusive results were obtained, highlighting the need for further research. 
Future investigations should prioritize tasks involving temporal reasoning to address biases in GPT-3.5 and GPT-4, considering diverse datasets and prompt structures.

\section{Limitations}
Our study's findings, drawn from the analysis of GPT-3.5 and GPT-4, suggest that the identified patterns may be specific to these models and not universally applicable to language models with different architectures or training methodologies. Given the continuous development of language models and the potential for new versions with updates, the biases observed in GPT-3.5 and GPT-4 may not persist in future releases. Recognizing the impact of prompt types on model performance, our study emphasizes the ongoing need for exploration to determine the most effective prompt types across different contexts. While the QA prompt showed improved predictions in some cases, the Textual Entailment format proved beneficial in others, underscoring the importance of selecting appropriate prompt types for comprehensive analyses. Interestingly, the ``BEFORE''/``AFTER'' bias observed in the QA format and TE consistent pair implicit events did not reappear in TE consistent pair explicit events, potentially influenced by the lower percentage of data in this category.

% 


\bibliography{custom}

%\nocite{jung-etal-2011-building,Tbcorpus,10.1007/s11280-017-0501-x,future-event,choubey-huang-2017-sequential,zhang-etal-2022-extracting,vashishtha-etal-2020-temporal,OpenAI,verhagen-etal-2010-Tempeval2,ning2018multi,zhou2019going,zhou2020temporal,zhou2021temporal,ning-etal-2018-cogcomptime,han-etal-2019-joint,wang-etal-2020-joint,zhang-etal-2022-extracting, kolomiyets-etal-2012-extracting,4338327}
\appendix

\clearpage

%\titlespacing*{\subsection} {0pt}{0ex}{0ex}
\onecolumn
\section{Appendix}
\label{sec:appendix}

\subsection{Experimental Details}
Below are few with the experimental results and other data gathered from our experiments in both implicit and explicit event for both Textual Entailment and Question Answering format.
\subsubsection{Textual Entailment}
\begin{figure*}[!ht]
\textbf{\begin{center}\small
   Model's Consistency 
\end{center} }\par\medskip
\begin{minipage}[b]{0.5\linewidth}
\includegraphics[width=8cm]{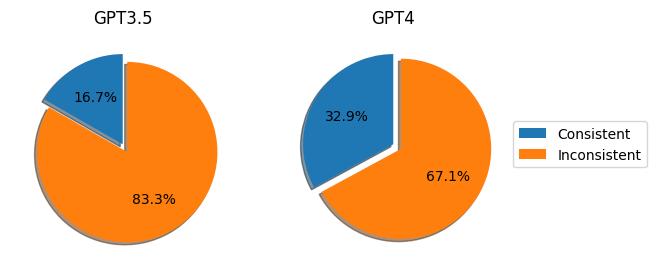}
\caption{\small Consistency in Response for implicit events}\label{label-a}
\end{minipage}\qquad
\begin{minipage}[b]{0.5\linewidth}
\includegraphics[width=8cm]{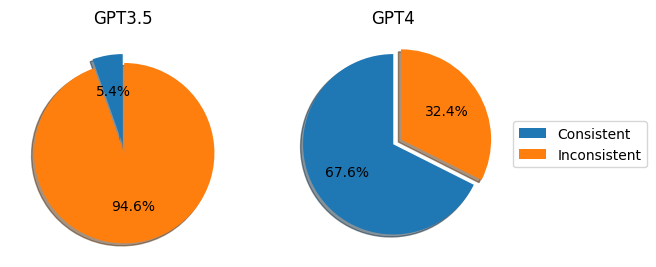}
\small\caption{\small Consistency in Response for explicit events}\label{label-b}
\end{minipage}
 \parbox{\textwidth}{
 \vspace{10pt}
Although our anticipation was for the model to provide one TRUE and one FALSE for each pair, we observed a discrepancy. The models did yield (``TRUE'',``FALSE'') for certain instances, but surprisingly, they often produced (``TRUE'',``TRUE'') or (``FALSE'',``FALSE''). Figures 7 and 8 visually depict the inconsistency discussed in Section \ref{sec:results}. Notably, we observe that GPT-4 exhibits more consistency than GPT-3.5 for both implicit and explicit events when prompts are presented in the Textual Entailment format.}
\end{figure*}

% % \subsection{Consistency in the model}

% % ---------------------------------------------------
% \subsection{BEFORE/AFTER Bias in Consistent pair (Implicit Events)}
\begin{figure*}[!ht]
\textbf{\begin{center}\small
  BEFORE/AFTER Bias in Consistent pair (Implicit Events)
\end{center} }\par\medskip
\begin{minipage}[b]{0.5\linewidth}
\includegraphics[width=8cm]{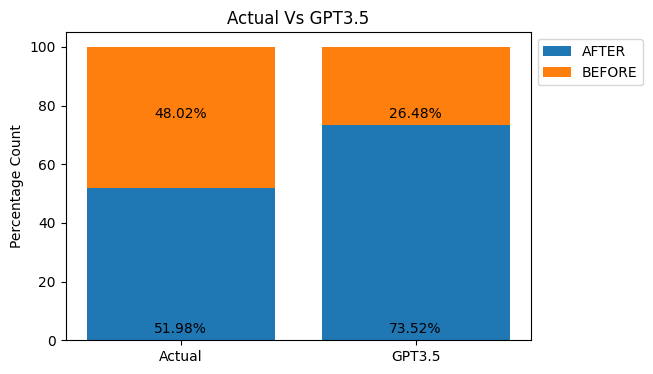}
\caption{\small GPT-3.5 biased towards AFTER }\label{label-te-gpt35-implicit}
\end{minipage}\qquad
\begin{minipage}[b]{0.5\linewidth}
\includegraphics[width=8cm]{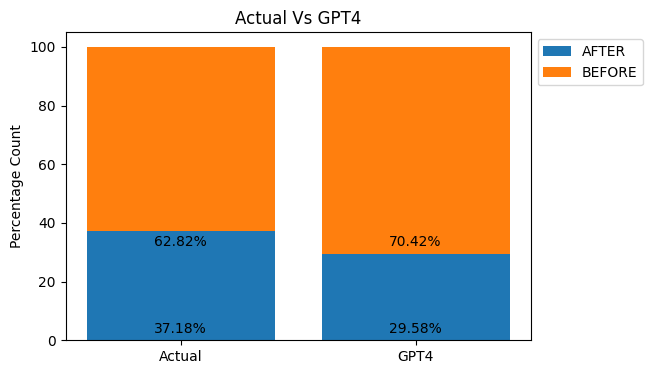}
\small\caption{\small GPT-4 biased towards BEFORE}\label{label-te-gpt4-implicit}
\end{minipage}
 \end{figure*}
\parbox{\textwidth}{
 \vspace{2pt}
 Figures 9 \& 10 visually illustrate the observed bias in the Textual Entailment consistent pair. As outlined in the Section \ref{sec:results}, we note GPT-3.5 demonstrating a bias towards ``AFTER'' and GPT-4 exhibiting a bias towards ``BEFORE''. While this behavior was previously observed in the Question Answering format for both implicit and explicit events, it is notable that in the Textual Entailment consistent format, this bias is observed only for implicit events.}
% % ------------------------------------------------
% % \subsection{Unbiased Consistent pair (Explicit Events)}
\onecolumn
\vspace{0pt}
\begin{figure*}[!ht]
\textbf{\begin{center}
  Unbiased consistent pair (Explicit Events)
\end{center} }\par\medskip
% \textbf{Bias in Inconsistent pair}\par\medskip
\begin{minipage}[b]{0.5\linewidth}
\vspace{0pt}
\includegraphics[width=8cm]{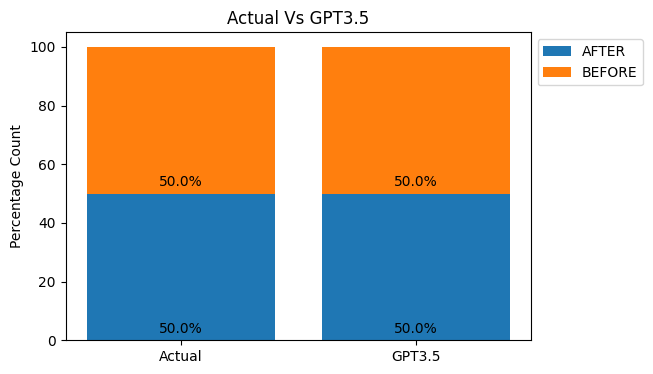}
\caption{\small Unbiased GPT-3.5}\label{label-te-gpt35-explicit}
\end{minipage}\qquad
\begin{minipage}[b]{0.5\linewidth}
\includegraphics[width=8cm]{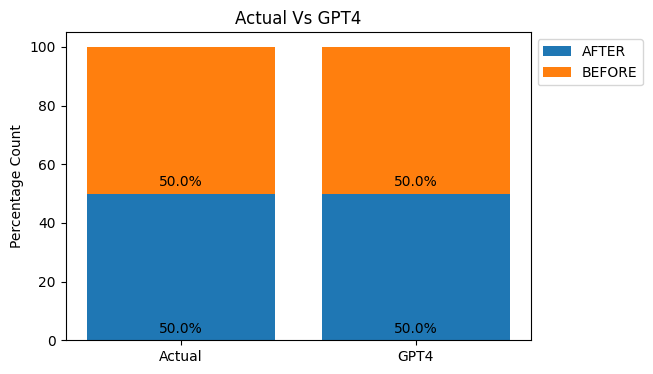}
\small\caption{\small Unbiased GPT-4}\label{label-te-gpt4-explicit}
\end{minipage}
 \parbox{\textwidth}{
 \vspace{0pt}
 In the preceding section for Consistent pair Textual Entailment, we noted that bias is observed in implicit events, attributed to their complexity. However, explicit events are better comprehended by the model, and the bias is absent here as shown in the  Figure 11 \& 12 .\\}
\end{figure*}
%%-------------------------------------------------------------
% % \subsection{Results of TE(Inconsistent pair)}
%%-------------------------------------------------------------
\begin{table*}[!htp]
\centering
\textbf{\begin{center}
  Results of TE(Inconsistent pair)
\end{center} }\par\medskip
  \fontsize{10pt}{10pt}\selectfont
  \begin{tabular}{|c|c|c|c|c|c|}
    \hline
    \rule{0pt}{8pt}
   \textbf{Event} & \textbf{Relation} & \textbf{Actual} & \textbf{GPT-3.5} & \textbf{GPT-4} \\
    \hline
    \rule{0pt}{8pt}
   \multirow{2}{*}{\textbf{Implicit}} & TRUE & 50.0\% &\textcolor{blue}{75.07\%} & {7.83\%}\\
    & FALSE & 50.0\% & 24.93\% & \textcolor{blue}{92.17\%}\\
    \cline{1-5} % Add hline for the first three columns
     \rule{0pt}{8pt}
    \multirow{2}{*}{\textbf{Explicit}}& TRUE & 50.0\% & \textcolor{blue}{99.12\%} & 0.88\% \\
    & FALSE & 50.0\% & 36.15\%& \textcolor{blue}{58.94\%} \\
    \hline
  \end{tabular}
  \caption{Actual vs Predicted distribution of GPT-3.5 \& GPT-4 in TE.}
   \parbox{\textwidth}{
 \vspace{10pt}
Table 2 clearly shows that GPT-3.5 tends to favor "FALSE" for both implicit and explicit events, whereas GPT-4 shows a preference for "TRUE" in both event types.\\}
  \label{tab:spanned-columns-te}
\end{table*}

% % ------------------------------------------------
% %\onecolumn
% % \subsection{Results of QA}
\begin{table*}[!htp]
\centering
\subsubsection{Question Answering}
\textbf{\begin{center}
  Results of QA
\end{center} }\par\medskip
  \fontsize{10pt}{10pt}\selectfont
  \begin{tabular}{|c|c|c|c|c|c|}
    \hline
    \rule{0pt}{8pt}
   \textbf{Event} & \textbf{Relation} & \textbf{Actual} & \textbf{GPT-3.5} & \textbf{GPT-4} \\
    \hline
    \rule{0pt}{8pt}
   \multirow{2}{*}{\textbf{Implicit}} & BEFORE & 57.8\% & 45.73\% &\textcolor{blue}{58.45\%}\\
    & AFTER & 42.2\% & \textcolor{blue}{53.48\%} & 39.43\%\\
    \cline{1-5} % Add hline for the first three columns
     \rule{0pt}{8pt}
    \multirow{2}{*}{\textbf{Explicit}}&BEFORE & 57.99\% & 48.29\% & \textcolor{blue}{67.07\%} \\
    & AFTER & 42.01\% & \textcolor{blue}{51.71}\%& 32.93\% \\
    \hline
  \end{tabular}
  % \quad
  \caption{Actual vs Predicted distribution of GPT-3.5 \& GPT-4 in QA}
  \label{tab:spanned-columns-qa}
   \parbox{\textwidth}{
 \vspace{10pt}
As previously discussed in Section \ref{sec:results}, Table 3 shows GPT-3.5 demonstrates a bias toward the ``BEFORE'' relation for both implicit and explicit events. Conversely, GPT-4 exhibits a conflicting bias, showing a preference for the ``AFTER'' relation in both types of events.\\}
\end{table*}
% % ------------------------------------------------
% % \subsection{Results of TE(Inconsistent pair)}
% % -----------------------------------------------
\subsection{Additional Results}
% \subsubsection{Textual Entailment}
\begin{figure*}[!ht]
\textbf{\begin{center}\small
   Consistent TE GPT-3.5
\end{center} }\par\medskip
\begin{minipage}[b]{0.5\linewidth}
\includegraphics[width=8cm]{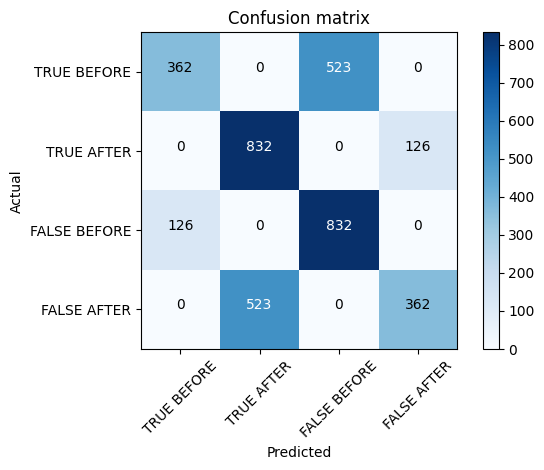}
\caption{\small Confusion matrix of GPT-3.5 for implicit events}\label{label-cfm_consisET_implicit_gpt3_5}
\end{minipage}\qquad
\begin{minipage}[b]{0.5\linewidth}
\includegraphics[width=8cm]{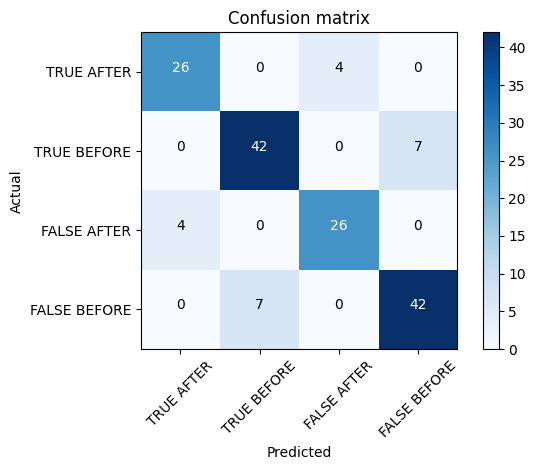}
\small\caption{\small Confusion matrix of GPT-3.5 for explicit events}\label{label-cfm_consisET_implicit_gpt4}
\end{minipage}
 \parbox{\textwidth}{
 \vspace{10pt}
Figure 13 \& 14 display the confusion matrix of GPT-3.5 for TE consistent pairs. These images vividly illustrate the breakdown of actual and predicted values for the BEFORE and AFTER relations. As mentioned earlier in Section \ref{sec:results}, We see that GPT-3.5 tends to exhibit bias towards AFTER in implicit events, while remaining unbiased for explicit events. }
\end{figure*}

% % \subsection{Consistency in the model}

% % ---------------------------------------------------
% \subsection{BEFORE/AFTER Bias in Consistent pair (Implicit Events)}
\begin{figure*}[!ht]
\textbf{\begin{center}\small
  Consistent TE GPT-4
\end{center} }\par\medskip
\begin{minipage}[b]{0.5\linewidth}
\includegraphics[width=8cm]{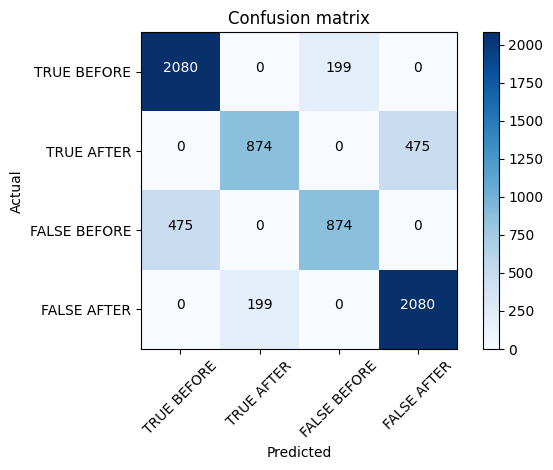}
\caption{\small Confusion matrix of GPT-4 for implicit events }\label{label-cfm-te-gpt35-implicit}
\end{minipage}\qquad
\begin{minipage}[b]{0.5\linewidth}
\includegraphics[width=8cm]{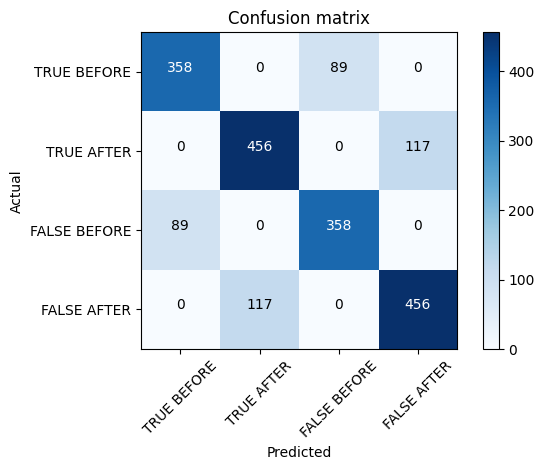}
\small\caption{\small Confusion matrix of GPT-4 for explicit events.}\label{label-cfm-te-gpt4-implicit}
\end{minipage}

\parbox{\textwidth}{
 \vspace{2pt}
Figures 15 and 16 present the confusion matrix of GPT-4 for TE consistent pairs. Once more, they depict the breakdown of actual and predicted values for the BEFORE and AFTER relations. As previously discussed in Section \ref{sec:results}, we observe that GPT-4 demonstrates a tendency towards biasing predictions towards BEFORE in implicit events, which contrasts with the behavior of GPT-3.5. However, GPT-4 remains unbiased for explicit events.}
 \end{figure*}

\end{document}